\documentclass[review]{elsarticle}
\usepackage{lineno,hyperref}
\modulolinenumbers[5]

\journal{Journal of Computer Speech and Language}

\usepackage{url}
\usepackage{latexsym}
\usepackage{graphicx}
\usepackage{color}
\usepackage{booktabs}
\usepackage{subcaption}
\usepackage{multirow}
\usepackage{amsmath}
\usepackage{placeins}
\usepackage{amssymb}
\usepackage[normalem]{ulem}
\usepackage{comment}
\usepackage{color}
\usepackage{colortbl}
\usepackage{tcolorbox}

\begin{document}

\begin{frontmatter}

\title{On the Effect of Dropping Layers of Pre-trained Transformer Models}


% \author{Hassan Sajjad, Fahim Dalvi, Nadir Durrani, \hbox{\rm and} Preslav Nakov \\
% Qatar Computing Research Institute --- HBKU, Doha, Qatar \\
%   {\tt \{hsajjad,faimaduddin,ndurrani,pnakov\}@hbku.edu.qa}
%  } 

\author{
 Hassan Sajjad$^{\clubsuit}$\footnote[1]{The work was done while the author was at QCRI} \hspace{2mm} Fahim Dalvi$^{\diamondsuit}$ \hspace{2mm} Nadir Durrani$^{\diamondsuit}$ \hspace{2mm} Preslav Nakov$^{\spadesuit}$$^1$ \\
 $^\clubsuit$Faculty of Computer Science, Dalhousie University, Canada \\
 $^{\diamondsuit}$Qatar Computing Research Institute, Hamad Bin Khalifa University, Qatar \\
 $^{\spadesuit}$Mohamed bin Zayed University of Artificial Intelligence, Abu Dhabi, UAE\\
 {\sf\small hsajjad@dal.ca,\{faimaduddin, ndurrani\}@hbku.edu.qa, preslav.nakov@mbzuai.ac.ae}
}

% \author[dal]{Hassan Sajjad}
% \author[hbku]{Fahim Dalvi}
% \author[hbku]{Nadir Durrani}
% \author[mb]{Preslav Nakov}
% \address[dal]{Faculty of Computer Science, Dalhousie University, Canada}
% \address[hbku]{Qatar Computing Research Institute, Hamad Bin Khalifa University, Qatar}
% \address[mb]{Mohamed bin Zayed University of Artificial Intelligence, Abu Dhabi, UAE}

\begin{abstract}

Transformer-based NLP models are trained using %hundreds of millions or even 
billions of parameters, limiting their applicability in computationally constrained environments.  While the number of parameters generally correlates with performance, it is not clear whether the entire network is required for a downstream task. Motivated by the recent work on pruning and distilling pre-trained models, we explore strategies to drop layers in pre-trained models, and observe the effect of pruning on downstream GLUE tasks. We were able to prune BERT, RoBERTa and XLNet models up to 40\%, while maintaining up to 98\% of their original performance. Additionally we show that our pruned models are at par with those built using knowledge distillation, both in terms of size and performance. Our experiments yield interesting observations such as: (i) the lower layers are most critical to maintain downstream task performance, (ii) some tasks such as paraphrase detection and sentence similarity are more robust to the dropping of layers, and (iii) models trained using different objective function 
exhibit different learning patterns and w.r.t the layer dropping.\footnote{The code is available at \url{https://github.com/hsajjad/transformers/}.} 

\end{abstract}


\end{frontmatter}

%\linenumbers

\section{Introduction}

Pre-trained Transformer models have achieved state-of-the-art performance on natural language processing tasks and have been adopted as feature extractors for solving downstream tasks such as question answering, natural language inference, and sentiment analysis. The current state-of-the-art Transformer-based pre-trained models consist of dozens of layers and millions of parameters. While deeper and wider models yield better performance, they also need large GPU/TPU memory. For example, BERT-large~\cite{devlin-etal-2019-bert} is  
trained with 335 million parameters, and requires at least 24 GB of GPU memory to load.
The larger size of these models limits their applicability in time- and memory-constrained environments.% such as small hand-held devices.

Several methods have been proposed to reduce the size of pre-trained models. Notable approaches include pruning parts of the network after training~\cite{NIPS2019_9551, voita-etal-2019-analyzing, mccarley2019pruning}, reduction through weight factorization and sharing~\cite{lan2019albert}, compression via knowledge-distillation~\cite{sanh2019distilbert} and quantization~\cite{zafrir2019q8bert, shen2019qbert}. Our work falls under the class of pruning methods.

The central argument governing pruning methods is that %Transformer 
deep neural models are over-parameterized and that not all parameters are strictly needed, especially at the inference time.
For example, previous research has 
%have 
shown that most of the attention heads can be removed \cite{heads-nips2019,voita-etal-2019-analyzing} or reallocated \cite{peng2020mixture} %at test time 
without significantly impacting performance. \citet{Gordon2019CompressingBS} pruned the least important weights in the network. We build our work based on similar observations, but we are interested in (\emph{i})~whether it is necessary to use all layers of a pre-trained model for downstream tasks, and if not, (\emph{ii})~which layers are necessary to keep in order to maintain good task-specific performance while achieving efficiency in transfer learning.

Motivated by recent findings in representation learning, we propose novel strategies to drop layers in pre-trained models. \citet{voita-etal-2019-bottom} showed that the top layers are biased towards the pre-training objective, leading us to question whether they are necessary for downstream tasks. \citet{heads-nips2019,dalvi-etal-2020-analyzing} discussed over-parameterization and the redundancy in pre-trained models, leading us to question whether adjacent layers contain redundant information. More concretely, we drop top, %layers, 
bottom, %layers, 
middle, %layers, 
or alternate layers in the network. 
We additionally present methods to find layers that contribute least in the network by using their activation patterns and weights. % of each layer atowards a downstream task and drop these.
%\nd{Give more details what these methods are using} %or layers that contribute least in the network. 
We apply our strategies on four state-of-the-art pre-trained models, BERT~\cite{devlin-etal-2019-bert},  RoBERTa~\cite{roberta}, ALBERT~\cite{lan2019albert} and XLNet~\cite{yang2019xlnet}. The %former two 
first three are auto-encoders, while XLNet is an auto-regressive model. %while 
ALBERT presents an interesting case in the mix as its layers share parameters.
%is a parameter sharing model. 
We additionally experiment using DistilBERT to analyze whether a distilled model can be pruned further. We evaluate against GLUE benchmark~\cite{wang-etal-2018-glue} %which involves a variety of
a suite of language understanding tasks. 
%We first remove a set of layers from the pre-trained model based on a strategy, and we then fine-tune the reduced model towards downstream tasks. We also compare our results to distilled versions of BERT and RoBERTa. 
Our findings are summarized below: %as follows:

%Our contributions and findings are the follows:

%\textcolor{red}{mix findings with analysis of task and distilled model}
\begin{itemize}
    \item We propose practical strategies to drop %a set of 
    layers in pre-trained models for efficient transfer learning.
    \item We show that dropping top layers works consistently well across different tasks and pre-trained models, e.g.,~yielding 40\% reduction in size while preserving up to 98.2\% of the performance.
    \item Our reduced models perform on par with models built using knowledge distillation in terms of accuracy, model size and inference speed, without requiring costly training of a new model.
    \item One-third of a distilled models can also be pruned successfully with an average loss of 0.75 points 
    \item %ALBERT, 
    Despite having cross-layer parameter sharing, ALBERT can still be pruned for efficient inference with a small drop in performance.
    \item Certain downstream tasks require as few 3 layers to maintain performance within 1\% threshold.
    \item Comparing architectures, models show different learning dynamics. For example, compared to BERT, RoBERTa and XLNet learn task-specific knowledge earlier in the network and are thus more robust to layer-dropping.
    %\item Our findings support selective distillation, i.e.,~aggressive distillation to the top layers and preserving the bottom layers of the pre-trained model.
\end{itemize}

%The remainder of this paper is organized as follows:  Section~\ref{sec:relatedwork} summarizes the related work. Section~\ref{sec:strategies} presents our layer-dropping strategies. Section~\ref{sec:experiemnts} describes our experimental setup. Section~\ref{sec:results} presents the results. Section~\ref{sec:discussion} offers an in-depth discussion. Finally, Section~\ref{sec:conclusion} concludes. % and points to possible directions for future work.
\paragraph{Contribution}
While a number of studies partially overlap with the strategies and the findings presented in this work, 
%we would like to emphasize that 
this is the first work that thoroughly investigates the effect of various layer-dropping methods using a variety of pre-trained models and on a large number of tasks. 
%Additionally, the previous work is limited to one model i.e. BERT. 
We showed that i) models have different learning dynamics, %and 
%Moreover, %an optimal 
ii) a smaller close to optimal network can be achieved by optimizing the number of layers to drop with respect to the task at hand, %We additionally show that 
iii) a distilled model can also benefit from layer-dropping. 
Our work recommends to use top layer-dropping as an essential baseline when building distilled models. Moreover, it provides a cheap way to get smaller models of any architecture rapidly, that are both memory and speed efficient.

\section{Related Work}
\label{sec:relatedwork}

\textbf{Efficient Pre-trained Models:}
% Recent %research 
% work has explored various methods to down-scale pre-trained models. They
Work done on exploring methods to down-scale pre-trained models can be categorized into architecture-invariant compression~\cite{lan2019albert,cao2019,shen2019qbert}, %quantization~\cite{zafrir2019q8bert,shen2019qbert}, 
knowledge distillation~\cite{Hinton06,sanh2019distilbert}, and pruning~\cite{Gordon2019CompressingBS,Guyon:jmlr}. 

Quantization~\cite{zafrir2019q8bert,shen2019qbert}, an \textit{architecture-invariant} method, reduces the numerical precision of the weights of the model to fewer bits. % , thus reducing the model size and limiting computation time. %Another architecture-invariant method is weight sharing across layers.~\citet{lan2019albert} proposed a lighter BERT that shares weight matrices across the hidden layers of the model, thus reducing the multi-layer parameters to single-layer parameters. 
\textit{Knowledge distillation (KD)} also known as student-teacher model \cite{hinton2015distilling}
%aims to 
trains a smaller model
%, called a \emph{student model}, 
that mimics the behavior of the larger model.
%\emph{teacher model}. 
Researchers have experimented with learning from the outputs of the encoder layers~\cite{sun-etal-2019-patient,jiao2019tinybert}, from the output logits~\cite{sanh2019distilbert,tang2019distiltaskspecific}, and from the attention maps~\cite{jiao2019tinybert,mobileBERT}. Another distinction is between general-purpose distilled models~\cite{sanh2019distilbert,mobileBERT} and task-specific ones~\cite{jiao2019tinybert,turc2019wellread,tang2019distiltaskspecific,sun-etal-2019-patient,tsai2019small}. 

\textit{Pruning} methods involve removing some parts of the networks that are either redundant or less relevant to the task at hand. \cite{Gordon2019CompressingBS,Guyon:jmlr,Renda2020ComparingRA}~pruned the least important weights in the network. \citet{heads-nips2019,voita-etal-2019-analyzing} demonstrated that %several of 
most of the attention heads can be pruned at test time, which reduces the computation, and speeds up inference. 
\citet{fan2019reducing} introduced \textit{LayerDrop} during training that resulted in pre-trained models that are robust towards dropping of layers at inference time. %and showed that their training method encourages the robustness of the model and facilitates dropping layers at inference time with minimal impact on performance. 
%\sout{Our work here focuses on pruning methods. However, unlike pruning individual weights and attention heads, we prune entire layers of the pre-trained neural network.} 
%The work of \citet{fan2019reducing}} \\textcolor{blue}{
%Their work is closest to ours as they also 
Our work is similar to them as we also remove layers from the network. %Different from them, 
But we show that layers can be dropped safely from the pre-trained models without the need for additional training using LayerDrop. Nevertheless our strategies can also be applied to a model trained using LayerDrop.

Recently, \citet{greedy_layer_pruning} proposed a greedy layer pruning method that drops layers based on their independent performance on the end task. Their assumption is that a local decision about a layer aligns with a globally correct selection of layers. We demonstrate that our results are comparable to theirs, but we need no additional training to find an optimal set of layers.

\citet{sun-etal-2019-patient,Xu2020BERTofTheseusCB} used the bottom six layers of the BERT-base model to initialize the student model. %As a baseline, they fine-tuned untrained student model on the GLUE tasks. 
This is %comparable 
similar to one of our strategies. However, their performance is much lower compared to our method. %\sout{than what our strategies offer.} 
Moreover, we provide a comprehensive evaluation of our strategies on four pre-trained models to prove their efficacy in reducing the size of the network. % while maintaining performance that is on par with several distilled models.

%\nd{Isn't the paragraph below conflicting with the work you have cited above? You already said Fan introduced LayerDrop}

%While pruning layers of a network is used in computer vision~\citet{Huang2016DeepNW}, ours is the first work to present a detailed empirical evidence and analysis of its effectiveness for NLP tasks, specifically on pre-trained models. We further show that our layer-dropping strategies are complementary to distillation methods and can be applied on top of distilled models. 

\citet{liu2020fastbert,schwartz2020right,xin2020deebert,zhou2020bert} speed up the inference time by introducing dynamic exiting strategies. The limitation of their work are the memory footprints of the model that remain identical to the original model. 
%suggested a dynamic early exiting strategy for accelerating BERT inference. Their 40% percent improvement is comparable to ours. However, they did not space an memory. 

\noindent\textbf{Representation analysis:} 
A number of studies have analyzed representations of pre-trained models at layer-level and showed that they learn
linguistic information~\cite{belinkov:2017:acl,dalvi-etal-2017-understanding,belinkov-etal-2017-evaluating,conneau2018you,liu-etal-2019-linguistic,tenney2019learn,tenney-etal-2019-bert,durrani-etal-2019-one,belinkov-etal-2020-linguistic,arps_syntax:arxiv22}. \citet{belinkov-etal-2020-interpretability,neuronSurvey} provided a comprehensive literature review of such work.
%A typical procedure to analyze representation is a post hoc analysis using a probing classifier. 
While the representation analysis uncovers, what linguistic properties different layers capture, %have learned learn about linguistic properties, it does 
they do not reflect which layers are important for transfer learning to a downstream task. Recently, \citet{Tamkin2020InvestigatingTI,Merchant2020WhatHT,durrani-etal-2021-transfer} %and \citet{Merchant2020WhatHT} 
attempted to address this by analyzing layer-wise transferability of features during fine-tuning. \citet{Tamkin2020InvestigatingTI} reinitialized individual layers of pre-trained model and observed the effect on the fine-tuning performance. \citet{Merchant2020WhatHT} used %three methods --
probing classifier, layer-wise similarity and layer-ablation for their analysis. Our work is similar to their layer-ablation study %where it is mainly 
which they %introduced 
carried out to understand the difficulty of a downstream task, but the premise of our work is very different. We %aim to 
gauge the importance of various subsets of layers with respect to the performance on downstream tasks, %in order 
to achieve efficient models. \citet{durrani-etal-2021-transfer} used layer-wise and neuron probing classifiers \cite{dalvi:2019:AAAI,durrani-etal-2020-analyzing} and showed that core-linguistic knowledge is preserved in the lower layers of fine-tuned models. This resonates with our empirical finding that shows that higher layers can be safely pruned for efficient transfer learning.

\section{Methodology}
\label{sec:strategies}

 \begin{figure}[t]
     \centering
     \includegraphics[width=\linewidth]{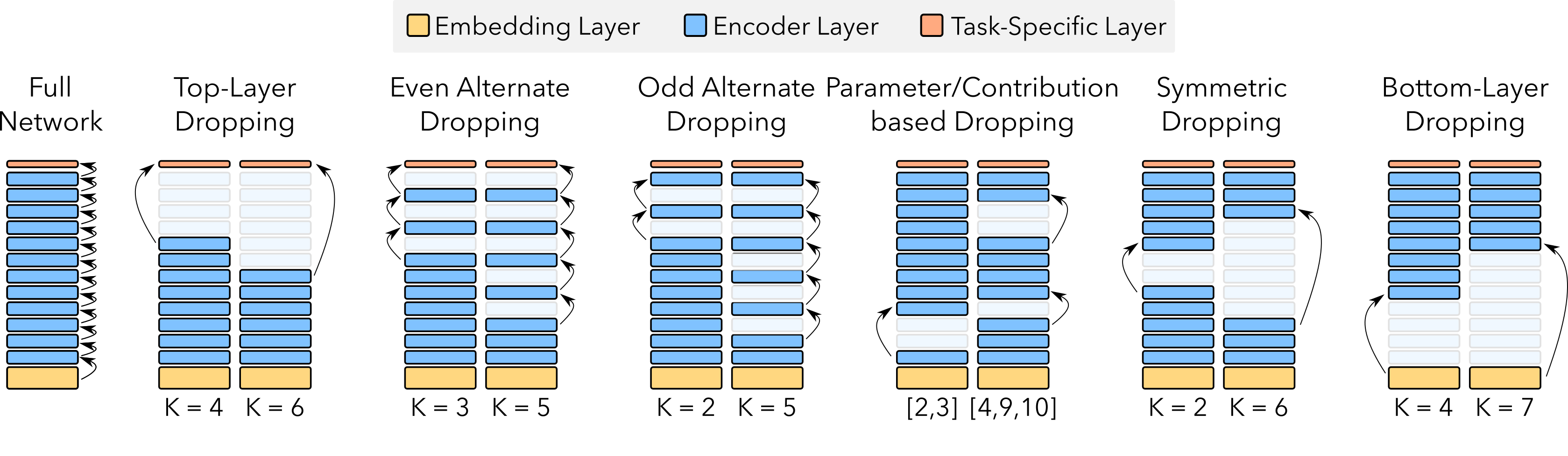}
     \caption{Illustration of layer-dropping strategies. $K$ represents the number of layers that are dropped. For example, $K=4$ in the top-layer strategy means top four layers of the model are dropped. In the contribution-based dropping, we select layers based on a similarity threshold. The number mentioned in the figure e.g. [2,3] shows the layers which are dropped based on the similarity threshold.}
     \label{fig:approach}
 \end{figure}

Consider a pre-trained language model $\mathbf{M}$ with an embedding layer $E_0$ and $L$ encoder layers: $\{l_1, l_2,\ldots, l_L\}$.  
We probe whether it is necessary to keep all layers of the network for downstream tasks.
We explore six strategies, that we describe below (also shown in Figure \ref{fig:approach}), to drop encoder layers from the model. %$\mathbf{M}$ as shown in Figure \ref{fig:approach}, 
Each pruning regime is 
followed by task-specific fine-tuning to analyze the effect of layer-dropping on the performance of the task. 

\subsection{Top-Layer Dropping}
The top layers in pre-trained models are specialized towards the underlying objective function \cite{voita-etal-2019-bottom}. \citet{zhang2020revisiting} reinitialized the upper layers when fine-tuning towards GLUE task. We hypothesize that %these 
the top layers may not be important when fine-tuning towards the %objective of 
a downstream task.
In this strategy, we drop top $K$ layers from the model. The output of layer $l_{L-K}$ serves as the last layer of the reduced network. Then, a task-specific layer is added on top of this layer to perform task-specific fine-tuning. 
Figure \ref{fig:approach} shows an example with dropping top 4 and 6 layers.

\subsection{Alternate Dropping}
Deep neural networks are %characterized by redundancy of information across the network 
innately redundant. %\cite{tang2019distiltaskspecific}.
~\citet{sun-etal-2019-patient} and \citet{jiao2019tinybert} amalgamated information from adjacent layers of the teacher model into %one 
a single layer of %a 
the student model.
We hypothesize that neighbouring layers preserve
similar information and may be dropped safely without any substantial loss of information.
%\nd{MERGE PARAGRAPH}
%In this strategy, we 
We drop $N$ alternating odd or even layers from top to bottom of the network. For example %for
in a 12-layer model with $K=4$, we consider two sets of alternate layers: \emph{Odd-alternate Dropping} -- \{5,7,9,11\} and \emph{Even-alternate Dropping} -- \{6,8,10,12\}, see Figure \ref{fig:approach} for illustration. 
When dropping an in-between layer $l_i$, the output of the previous layer $l_{i-1}$ %would 
becomes the input of the next layer $l_{i+1}$, %Note that dropping $l_i$ would 
causing a mismatch in the expected input to $l_{i+1}$. However, we %assume 
hope that during task-specific fine-tuning, the model will recover from this discrepancy. 

\subsection{Parameter-Based Dropping}
In this approach, we estimate the importance of a given layer based on the model parameters. More specifically, we rank the layers based on their weights. We tested two hypotheses: 
(\emph{i})~higher \emph{magnitude} of the weights signals higher layer importance, (\emph{ii})~higher \emph{variance} of the weights corresponds to higher layer importance. We refer to the former as \texttt{Aggregation Method}, where we aggregate the weights of a layer, and we call the latter a \texttt{Variance Method}, where we calculate the variance of each layer. We drop the layers with the lowest aggregation or variance scores. Note that a transformer block has various sub-layers, but in our experiments we only used the final weights. We leave experiments with other layers within a transformer block as a possible direction for future work.

\subsection{Contribution-Based Dropping}
Our next strategy is based on the idea that a layer contributing below a certain threshold 
might be a good candidate for dropping. We define the contribution of a layer $l_i$ in terms of the cosine similarity between its input and its output representations.
A layer $l_i$ with a high similarity (above a certain threshold) indicates that its output has not changed much from its input, and therefore it can be dropped from the network.
More concretely, in the forward pass, we calculate the cosine similarity between the representation of the sentence token (CLS) before and after each layer. We average the similarity scores of each layer over the development set, and select layers that have an average similarity above a certain threshold for dropping. This \textit{contribution-based strategy} can be seen as a principled variation of \textit{alternate dropping}.

\subsection{Symmetric Dropping}

The bottom layers are closer to the input while the top layers are closer to the output. It is possible that both the top layers and the bottom layers are more important than the middle layers. %Thus, we further experiment with 
The \emph{Symmetric dropping} strategy %, where we keep 
retains the top and the bottom $X$ layers, and drop $K$ middle layers, where $2X+K = L$. For example, in a 12-layer model, if $K=6$, we retain three top and three bottom layers, dropping layers 4--9. The output of layer 3 would then serve as an input to layer 10. % as can be seen in Figure \ref{fig:approach}.

\subsection{Bottom-Layer Dropping}
Previous work on analyzing layers in Neural Networks~\cite{belinkov:2017:acl,tenney-etal-2019-bert, liu-etal-2019-linguistic,dalvi2022discovering,sajjad:naacl:2022} has shown that the lower layers model local interactions between words (which is important for morphology and lexical semantics), thus providing essential input to the higher layers. %Thus, 
Removing lower layers could be therefore catastrophic. We still perform these experiments %. %to verify what seems to be obvious 
for the sake of completeness. 
We remove the bottom $K$ layers of the model. The output of the embedding layer $l_0$ serves as an input to layer $l_{K+1}$ of the original model.

\section{Experimental Setup}
\label{sec:experiemnts}

\paragraph{Datasets}

We evaluated our strategies on General Language Understanding Evaluation (GLUE) tasks ~\citep{wang-etal-2018-glue}
%\footnote{See Appendix for data statistics and download link.} %It involves a variety of %language understanding 
tasks,  which serves as a defacto standard 
to evaluate pre-trained language models. Table \ref{tab:sentence_data_statistics} provides statistics of each dataset. More specifically, we evaluated on the following tasks: SST-2 for sentiment analysis with the Stanford sentiment treebank \cite{socher-etal-2013-recursive}, MNLI for natural language inference \cite{williams-etal-2018-broad}, QNLI for Question NLI \cite{rajpurkar-etal-2016-squad}, QQP for Quora Question Pairs,\footnote{\url{http://data.quora.com/First-Quora-Dataset-Release-Question-Pairs}} RTE for recognizing textual entailment \cite{Bentivogli09thefifth}, MRPC for Microsoft Research paraphrase corpus \cite{dolan-brockett-2005-automatically}, and STS-B for the semantic textual similarity benchmark \cite{cer-etal-2017-semeval}.
We left out WNLI, due to the irregularities in its dataset, as also reported by others,\footnote{\url{http://gluebenchmark.com/faq}} as well as CoLA due to large variance and unstable results across fine-tuning runs.

\begin{table}[]                                    
\centering                
\footnotesize
    \begin{tabular}{l|l|ll}                                    
        \toprule                                    
        Task  & Description & Train & Dev  \\        
        \midrule
        SST-2 & Sentiment analysis & 67349 &   872  \\
        MRPC  & Microsoft Research paraphrase corpus & 3668 &   408  \\
        MNLI  & Natural language inference & 392702 &  9815  \\
        QNLI  & Question natural language inference & 104743 &  5463  \\
        QQP   & Quora question pairs & 363846 & 40430  \\
        RTE   & Recognizing textual entailment & 2490 &   277  \\
        STS-B & Semantic textual similarity & 5749 &  1500 \\
    \bottomrule
    \end{tabular}
    \caption{Data statistics of the GLUE tasks.
    %Data statistics (number of sequences) on the official training and development sets used in the experiments. 
    All tasks are binary classification tasks, except for STS-B which is a regression task. Recall that the test sets are not publicly available, and hence we use development set to report results. 
    %10\% of the official train as development, and the official development set as our test set. Exact split information is provided in the code README. The data is available at \url{https://gluebenchmark.com/tasks.}
    }
\label{tab:sentence_data_statistics}                        
\end{table}

\paragraph{Models}
We experimented with three state-of-the-art 12-layered pre-trained models
\footnote{For the sake of clarity when the trends are similar across models, we present the results of selected models only.}
%, we move the results for a few models to the appendix.}  
BERT~\cite{devlin-etal-2019-bert}, RoBERTa~\cite{roberta} and XLNet~\cite{yang2019xlnet}. We additionally experimented using a 12-layered ALBERT~\cite{lan2019albert} model and a distilled model, DistilBERT~\cite{sanh2019distilbert}. Our selection of models encourage interesting comparison between different types of models such as auto-regressive vs. auto-encoder and a large model vs. its distilled version.  
All experiments are conducted using the \texttt{transformers} library~\cite{Wolf2019HuggingFacesTS}.
We used the default settings and did not optimize the parameters. %\footnote{We provide the parameter details with the code.}
%This caused a slight mismatch between some of the results reported in the original paper for each model and what we reported here.
We limit our experiments to the base versions of the transformers as we could not experiment
with BERT-large or XLNet-large due to memory limitations.\footnote{In order to fit large models in our TitanX 12GB GPU cards, we tried to reduce the batch size, but this yielded poor performance, 
see
%as previously reported by the BERT team  
\url{https://github.com/google-research/bert\#out-of-memory-issues}.}
However, our strategies are straightforward to apply to models of any depth. 

\paragraph{End-to-End Procedure}
Given a pre-trained model, we drop layers using one of the strategies described in Section~\ref{sec:strategies}. We then performed task-specific fine-tuning using GLUE training sets for three epochs as prescribed by \cite{devlin-etal-2019-bert}\footnote{We experimented with using more epochs, especially for dropping strategies that exclude in-between layers, in order to let the weight matrix adapt to the changes. However, we did not see any benefit in going beyond three epochs.} and evaluated on the official devsets. 
%three runs\footnote{\cite{devlin-etal-2019-bert} used \hl{10} runs and took the maximum performance run. We did not see much variation in several runs and we used three runs only.} \hl{lets add whats different in each run (initialization, seed?)} for each experiment (including the experiments with zero drop of layers) and we report the best result only. 
 %doing so, and thus %we 
%report the results %after
%using three epochs only.

\section{Evaluation Results}
\label{sec:results}
We experimented with dropping $K$ number of layers where $K=2,4,6$ in BERT, RoBERTa and XLNet, and $K=1,2,3$ in DistilBERT (a 6-layer model). As an example, for $K=2$ on a 12-layer model, we drop the following layers: top strategy -- $\{11,12\}$; bottom strategy -- $\{1,2\}$; even-alternate -- $\{10,12\}$; odd-alternate -- $\{9,11\}$; symmetric -- $\{6,7\}$. For the parameter-based strategy, we calculate the score of every layer based on the aggregated weights and the variance in the weights, and we drop the layers with the lowest score.
In the contribution-based strategy, the dropping of layers is dependent on a similarity threshold. We calculate the similarity between input and output of each layer and remove layers with similarity 
above the threshold values of $0.95$, $0.925$ and $0.9$. These values were chosen empirically. A threshold value below $0.9$ or above $0.95$ resulted in either more than half of the network being considered as similar, or none of the layers to be similar. 
%We discuss this approach further in Section~\ref{sec:discussion}.

\subsection{Comparing Strategies}
\begin{figure}[t]
    \begin{subfigure}[b]{\linewidth}
    \centering
    \includegraphics[width=\linewidth]{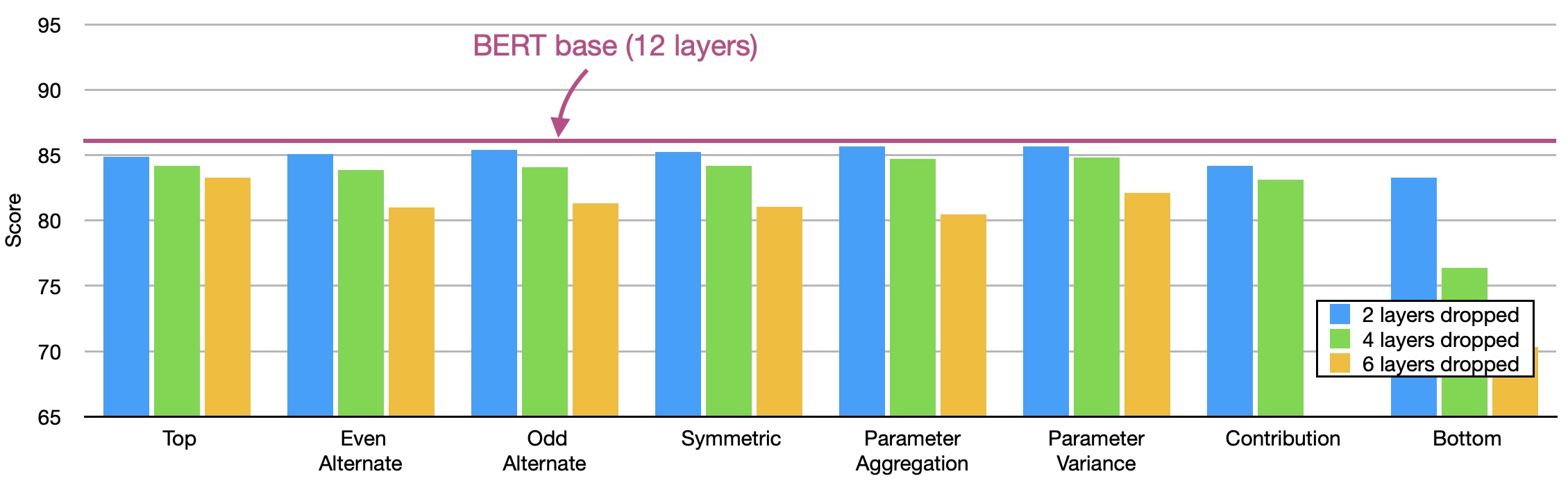}
    \caption{BERT}
    \label{fig:bert}
    \end{subfigure}
    \centering
    \begin{subfigure}[b]{\linewidth}
    \centering
    \includegraphics[width=\linewidth]{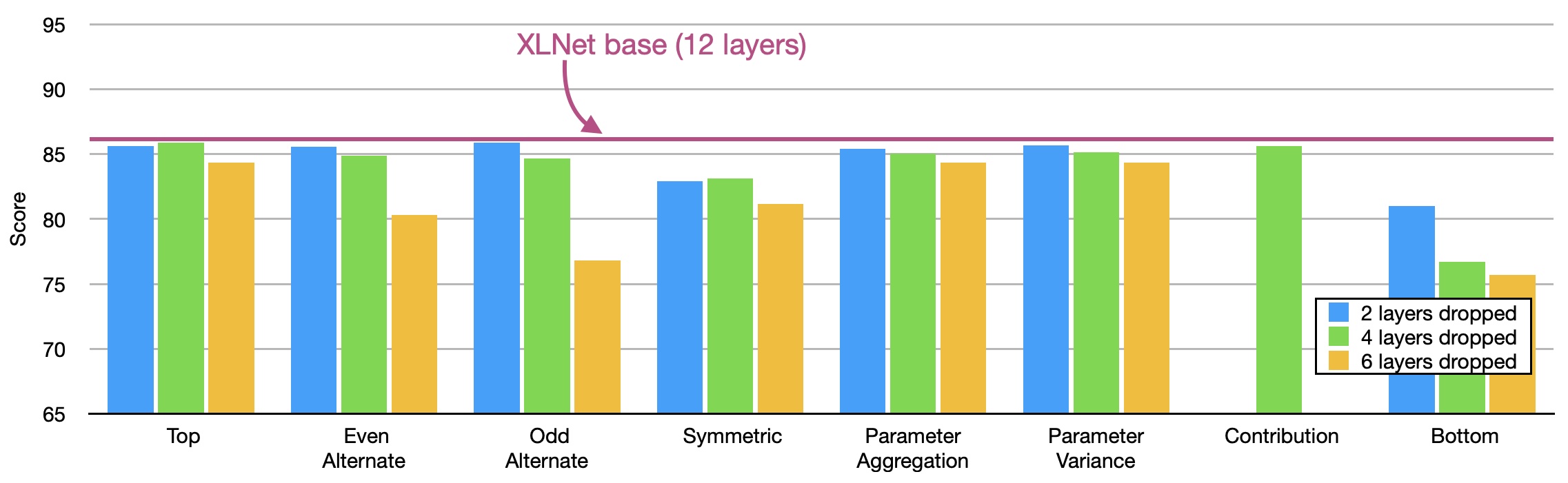}
    \caption{XLNet}
    \label{fig:xlnet}
    \end{subfigure}
    \caption{Average classification performance on GLUE tasks when using different layer-dropping strategies and when removing different numbers of layers for BERT and XLNet. Note that the contribution-based strategy selects layers based on the similarity threshold. In some cases it does not select (2,4 or 6) number of layers, which results in some missing bars in the figure. The horizontal red line represents the results using the full model.}
    \label{fig:average_results}
\end{figure}

Figure \ref{fig:average_results} presents average classification performance of BERT and XLNet using various layer-dropping strategies. We observe similar trends for RoBERTa and DistilBERT and limit the presentation of results to two models here.

\textbf{Top-layer dropping consistently outperforms other strategies when dropping 6 layers.} We dropped half of the top layers (yellow bars in the top strategy) with an average loss of only $2.91$ and $1.81$ points for BERT and XLNet respectively. %Similarly, dropping one-third of the network (i.e. $4$ layers) resulted in a drop of only $2$ points and $0.23$ points for BERT and XLNet respectively (blue bars in the top strategy). 
The \emph{Bottom-layer dropping} strategy performed the worst across all models, as expected, showing that it is more damaging to remove information from the lower layers of the network. The behavior of top and bottom dropping is consistent across all models. It nicely connects with findings in representation learning, i.e. lower layers learn core-linguistic phenomena and our results show that they are important to maintain task-specific performance.

\textbf{Parameter-based strategy using variance is the second best strategy at $K=6$.} Compared to most of the other strategies presented in this work, the parameter-based strategy makes a more informed decision based on the parameters of the model, i.e.,~the weights. We found the variance-based strategy to outperform the aggregation-based one, and thus we limit our discussion to the former only.
The variance-based method selected different layers to drop for each model. The order of the six layers to drop is $\{1,12,8,9,11,2\}$ for BERT, $\{11,12,6,7,5,10\}$ for RoBERTa and $\{11,12,7,8,9,10\}$ for XLNet. One common observation here is that the last 2--3 layers and the middle layers of the models can be removed safely with a small drop in performance (see the results of the variance-based method in Figure~\ref{tab:all_tasks}). Moreover, BERT is an exception where the first two contextualized layers $\{1,2\}$ are also selected to be removed.
%This could be due to the fact that BERT is more distributive and redundant~\cite{dalvi:featureselection:arxiv} and the knowledge . % even of its fundamental knowledge? So we saw that XLNet and RoBERTa when you FT them linguistic knowledge disappear from the higher layers but not in BERT. And I think due to this fact this layer removal happened in BERT. 
This resulted in a huge loss in performance (see the results for BERT when dropping 6 layers based on the variance-based method). Interestingly, dropping 6-layers of XLNet resulted in a model that was identical to that of the top-layer strategy, i.e.,~removing the top-6 layers. RoBERTa presents an interesting case where the parameter-based strategy resulted in a drop of the middle layers and of the top layers, while keeping the lower and the higher middle layers. The average results for RoBERTa when using the variance-based method are lower by 0.73 point only compared to the top-layer method. The promising results of the parameter-based method on two out of three models show its efficacy. Note that our current exploration is limited to the parameters of the base models. Fine-tuning substantially changes the parameters~\cite{durrani-etal-2021-transfer}, which may result in a task-wise informed dropping of layers. We did not try task-specific pruning as the focus of our work is on task-agnostic efficient models.

%Is this combination competitive to top-layer dropping? Table~\ref{tab:topvsvariance} compares the results. The parameter-based dropping did not out-perform the top-layer dropping on any task. However, the results are quite close with a maximum difference of 0.73 points only. }
%
%\textcolor{blue}{The results across three models conclude that the top-layer dropping is the most reliable strategy when dropping half of the network. The effectiveness of parameter-based strategy on two models suggests that it can be used as a principled approach to identify the importance of a layer. A task-specific layer selection after fine-tuning the model may further clarify the effectiveness of this strategy. However, our premise in this paper is to provide smaller models independent of the end task. }
%
%The results of variance-based strategy are quite competitive to the top-layer strategy e.g. on BERT, dropping 2 and 4 layers using the variance-based method resulted in better performance than the top-layer strategy. On XLNet, it is only 0.77 points lower than the top-layer strategy (85.91 vs. 85.14), when dropping 4 layers. %Interestingly when dropping the 6 layers, the variance-based method dropped identical layers of XLNet as that of using the top-layer strategy, thus resulted in a loss equal to the top-layer strategy. 

\textbf{Dropping top alternate layers is better than dropping top consecutive layers.}
The \emph{Odd-alternate dropping} strategy gave better results than the top at $K=2$ (blue bars in the Odd-alternate strategy), across all the tasks.
%At $K=2$ (blue bars in Figure \ref{fig:average_results}), both alternate dropping strategies showed better results than the top strategy whereas odd-alternate dropping outperformed all strategies across all models. 
Looking at the %exact 
layers that were dropped: top -- $\{11,12\}$; even-alternate -- $\{10,12\}$; odd-alternate -- $\{9,11\}$, we can say that (\emph{i})~dropping last two consecutive layers $\{11,12\}$ is more harmful than removing alternate layers, and (\emph{ii})~keeping the last layer $\{9,11\}$ is more important than keeping the second last layer with its alternate pair. 
%This is contrary to the general understanding that the last layer is well optimized for the pre-trained task and may not be essential for task-specific fine-tuning. 
%\textcolor{red}{We further analyzed the role of the last layer later in this section. --fd need to see if this matches discussion}
At $K=6$, the \emph{Alternate dropping} strategies show a large drop in the performance, perhaps due to removal of lower layers. Recall that our results from the bottom strategy showed %that 
lower layers %are 
to be critical for transfer learning.

The \emph{Symmetric strategy} gives importance to both top and bottom layers and drops the middle layers. Dropping two middle layers from BERT degrades the performance by $0.97$ points and makes it the second best strategy at $K=2$. 
However, on XLNet the performance degrades drastically when dropping the same set of layers. Comparing these two models, XLNet is sensitive to the dropping of middle layers while BERT shows competitive results to the \emph{Top-layer dropping} strategy even after removing 4 middle layers. We analyze the difference in the behavior of models in Section~\ref{sec:discussion}. % show competitive results to the best performing strategy, top dropping.
%\nd{I would either move the whole thing here or there. IMHO thing this XLNet paragraph does not belong here in the comparing strategy section}
% this is only one sentence. i left it here

For \emph{Contribution-based strategy}, we chose layers $\{3,5\}$ at threshold $0.95$ and $\{3,5,8,9\}$ at threshold $0.925$ for BERT, and layers $\{9,10,11\}$ at threshold $0.925$ and $\{8,9,10,11\}$ at threshold $0.9$ for XLNet. Using a lower or a higher similarity threshold resulted in dropping none or more than half of the layers in the network respectively. For BERT, the contribution-based dropping did not work well since the method chose a few lower layers for dropping.
On the contrary, it worked quite well on XLNet where higher layers were selected. This is in-line with the findings of top and bottom strategy that all models are robust to  dropping of higher layers compared to dropping of lower layers. 

The contribution-based strategy is based on the activations of each layer, which is an input-dependent process. Depending on the nature of the input or the task, the activation patterns will change. We suspect that this is one of the reasons for the failure of the strategy. A strategy based on task-specific contribution might yield a better performance. However, in this work we focused on task-independent efficient models, leaving task-dependent models for future work.

\begin{table}[t]									
\centering
\resizebox{1\textwidth}{!}{%
\begin{tabular}{l|lllll|ll}	

\toprule									
									
Drop. & SST-2 & MNLI & QNLI & QQP & STS-B  & RTE & MRPC \\		
\midrule						
%\multicolumn{8}{c}{BERTOLD} 			
%$0$ & 92.43 & 84.04 & 91.4 & 91 & 88.71  & 61.73 & 81.62
%$2$ & 91.74 ({\color{red}0.23$\downarrow$}) & 83.77 ({\color{red}0.27$\downarrow$}) & 89.16 ({\color{red}2.24$\downarrow$}) & 90.95 ({\color{red}0.05$\downarrow$}) & 88.24 ({\color{red}0.47$\downarrow$})  & 63.54 ({\color{blue}1.81$\uparrow$}) & 68.38 ({\color{red}13.24$\downarrow$})
%$4$ & 90.37 ({\color{red}1.60$\downarrow$}) & 83.16 ({\color{red}0.88$\downarrow$}) & 88.56 ({\color{red}2.84$\downarrow$}) & 90.79 ({\color{red}0.21$\downarrow$}) & 88.10 ({\color{red}0.61$\downarrow$})  & 62.45 ({\color{blue}0.72$\uparrow$}) & 80.88 ({\color{red}0.74$\downarrow$})
%$6$ & 90.37 ({\color{red}1.60$\downarrow$}) & 81.12 ({\color{red}2.92$\downarrow$}) & 87.70 ({\color{red}3.70$\downarrow$}) & 90.50 ({\color{red}0.50$\downarrow$}) & 88.16 ({\color{red}0.55$\downarrow$}) & 62.82 ({\color{blue}1.09$\uparrow$}) & 75.74 ({\color{red}5.88$\downarrow$})
%\midrule
\multicolumn{8}{c}{BERT} \\ 			
$0/12$ & 92.43 & 84.04 & 91.12 & 91.07 & 88.79  & 67.87 & 87.99 \\
$2/12$ & 92.20 ({\color{red}0.23$\downarrow$}) & 83.26 ({\color{red}0.78$\downarrow$}) & 89.84 ({\color{red}1.28$\downarrow$}) & 90.92 ({\color{red}0.15$\downarrow$}) & 88.70 ({\color{red}0.09$\downarrow$})  & 62.82 ({\color{red}5.05$\downarrow$}) & 86.27 ({\color{red}1.72$\downarrow$}) \\
$4/12$ & 90.60 ({\color{red}1.83$\downarrow$}) & 82.51 ({\color{red}1.53$\downarrow$}) & 89.68 ({\color{red}1.44$\downarrow$}) & 90.63 ({\color{red}0.44$\downarrow$}) & 88.64 ({\color{red}0.15$\downarrow$}) & 67.87 ({\color{blue}0.00}) & 79.41 ({\color{red}8.58$\downarrow$}) \\
$6/12$ & 90.25 ({\color{red}2.18$\downarrow$}) & 81.13 ({\color{red}2.91$\downarrow$}) & 87.63 ({\color{red}3.49$\downarrow$}) & 90.35 ({\color{red}0.72$\downarrow$}) & 88.45 ({\color{red}0.34$\downarrow$}) & 64.98 ({\color{red}2.89$\downarrow$}) & 80.15 ({\color{red}7.84$\downarrow$})  \\
\midrule
\multicolumn{8}{c}{RoBERTa} \\ 			
$0/12$ & 92.20 & 86.44 & 91.73 & 90.48 & 89.87 & 68.95 & 88.48 \\
$2/12$ & 93.46 ({\color{blue}1.26$\uparrow$}) & 86.53 ({\color{blue}0.09$\uparrow$}) & 91.23 ({\color{red}0.50$\downarrow$}) & 91.02 ({\color{blue}0.54$\uparrow$}) & 90.21 ({\color{blue}0.34$\uparrow$})  & 71.84 ({\color{blue}2.89$\uparrow$}) & 89.71 ({\color{blue}1.23$\uparrow$}) \\
$4/12$ & 93.00 ({\color{blue}0.80$\uparrow$}) & 86.20 ({\color{red}0.24$\downarrow$}) & 90.57 ({\color{red}1.16$\downarrow$}) & 91.12 ({\color{blue}0.64$\uparrow$}) & 89.77 ({\color{red}0.10$\downarrow$}) & 70.40 ({\color{blue}1.45$\uparrow$}) & 87.50 ({\color{red}0.98$\downarrow$}) \\
$6/12$ & 91.97 ({\color{red}0.23$\downarrow$}) & 84.44 ({\color{red}2.00$\downarrow$}) & 90.00 ({\color{red}1.73$\downarrow$}) & 90.91 ({\color{blue}0.43$\uparrow$}) & 88.92 ({\color{red}0.95$\downarrow$}) & 64.62 ({\color{red}4.33$\downarrow$}) & 85.78 ({\color{red}2.70$\downarrow$})  \\
\midrule
\multicolumn{8}{c}{XLNET}  \\				
$0/12$ & 93.92 & 85.97 & 90.35 & 90.55 & 88.01  & 65.70 & 88.48 \\
$2/12$ & 93.35 ({\color{red}0.57$\downarrow$}) & 85.67 ({\color{red}0.30$\downarrow$}) & 89.35 ({\color{red}1.00$\downarrow$}) & 90.69 ({\color{blue}0.14$\uparrow$}) & 87.59 ({\color{red}0.42$\downarrow$})  & 66.06 ({\color{blue}0.36$\uparrow$}) & 86.52 ({\color{red}1.96$\downarrow$}) \\
$4/12$ & 92.78 ({\color{red}1.14$\downarrow$}) & 85.46 ({\color{red}0.51$\downarrow$}) & 89.51 ({\color{red}0.84$\downarrow$}) & 90.75 ({\color{blue}0.20$\uparrow$}) & 87.74  ({\color{red}0.27$\downarrow$}) & 67.87 ({\color{blue}2.17$\uparrow$}) & 87.25 ({\color{red}1.23$\downarrow$}) \\
$6/12$ & 92.20 ({\color{red}1.72$\downarrow$}) & 83.48 ({\color{red}2.49$\downarrow$}) & 88.03 ({\color{red}2.32$\downarrow$}) & 90.62 ({\color{blue}0.07$\uparrow$}) & 87.45  ({\color{red}0.56$\downarrow$}) & 65.70 ({\color{blue}0.00}) & 82.84 ({\color{red}5.64$\downarrow$}) \\
\midrule
\multicolumn{8}{c}{DistilBERT}  \\		
$0/6$ & 90.37 & 81.78 & 88.98 & 90.40 & 87.14  & 60.29 & 85.05 \\
$1/6$ & 90.37 ({\color{blue}0.00}) & 80.41 ({\color{red}1.37$\downarrow$}) & 88.50 ({\color{red}0.48$\downarrow$}) & 90.33 ({\color{red}0.07$\downarrow$}) & 86.21 ({\color{red}0.93$\downarrow$})  & 59.93 ({\color{red}0.36$\downarrow$}) & 84.80 ({\color{red}0.25$\downarrow$}) \\
$2/6$ & 90.25 ({\color{red}0.12$\downarrow$}) & 79.41 ({\color{red}2.37$\downarrow$}) & 86.60 ({\color{red}2.38$\downarrow$}) & 90.19 ({\color{red}0.21$\downarrow$}) & 86.91 ({\color{red}0.23$\downarrow$})  & 62.82 ({\color{red}2.53$\uparrow$}) & 82.60 ({\color{red}2.45$\downarrow$}) \\
$3/6$ & 87.50 ({\color{red}2.87$\downarrow$}) & 77.07 ({\color{red}4.71$\downarrow$}) & 85.78 ({\color{red}3.20$\downarrow$}) & 89.59 ({\color{red}0.81$\downarrow$}) & 85.19 ({\color{red}1.95$\downarrow$}) & 58.48 ({\color{red}1.81$\downarrow$}) & 77.45 ({\color{red}7.60$\downarrow$}) \\
\bottomrule

% \multicolumn{8}{c}{ALBERT}  \\		
% $0/12$ & 89.79 & 83.39 & 90.24 & 90.29 & 89.61  & 67.51 & 74.02 \\
% $2/12$ & 91.40 ({\color{blue}1.61}) & 83.82 ({\color{blue}0.43$\downarrow$}) & 89.55 ({\color{red}0.69$\downarrow$}) & 89.64 ({\color{red}0.65$\downarrow$}) & 89.54 ({\color{red}0.07$\downarrow$})  & 69.31 ({\color{blue}1.80$\downarrow$}) & 88.48 ({\color{red}0.00$\downarrow$}) \\
% $4/12$ & 91.63 ({\color{blue}1.84$\downarrow$}) & 82.73 ({\color{red}0.66$\downarrow$}) & 90.24 ({\color{blue}0.00}) & 88.51 ({\color{red}1.78$\downarrow$}) & 87.00 ({\color{red}2.61$\downarrow$})  & 66.06 ({\color{blue}1.45$\uparrow$}) & 85.29 ({\color{red}0.00$\downarrow$}) \\
% $6/12$ & 90.14 ({\color{blue}0.35$\downarrow$}) & 81.64 ({\color{red}1.75$\downarrow$}) & 89.11 ({\color{red}1.13$\downarrow$}) & 90.08 ({\color{red}0.21$\downarrow$}) & 88.21 ({\color{red}1.40$\downarrow$}) & 53.79 ({\color{red}13.72$\downarrow$}) & 83.58 ({\color{red}0.00$\downarrow$}) \\
% \bottomrule

\end{tabular}
}
\caption{Task-wise performance for the top-layer dropping strategy using the official GLUE development sets. Drop. represents the number of layers that are dropped in comparison to the total number of layers in the model. The red numbers with downward arrow shows the drop in performance in comparison to using the full model i.e. 0/12 and the blue numbers with upward arrow shows the gain in performance.}
\label{tab:all_tasks}									
\end{table}

\subsection{Task-wise Results}
\emph{Top-layer strategy} works consistently well for all models at $K=6$. In the rest of the paper, we discuss the results for the \emph{Top-layer strategy} only, unless specified otherwise. 
Table~\ref{tab:all_tasks}\footnote{We use default settings provided in the Transformer library.
%to produce the results. 
This causes a slight mismatch between some numbers mentioned in the original papers of each models and our paper.} presents the results for 
the individual GLUE tasks using the \emph{Top-layer strategy} on three pre-trained models and a distilled model. We observe the same trend as for the averaged results: for most of the tasks, we can safely drop half of the top layers in BERT, RoBERTa and XLNet %at the cost of 
losing only 1-3 points.

\textbf{The paraphrase task (QQP) and sentence similarity task (STS-B) are least affected by the dropping of layers.}
%Note that the QQP and STS-B tasks are least affected by the dropping of layers across all models. 
When dropping half of the layers, %from the network, %12-layer models, 
there was no loss in performance for QQP on XLNet and RoBERTa, and %only 
a loss of $0.72$ only for BERT. Similarly, for STS-B we observed a decrease of only $0.56$, $0.95$ and $0.34$ points for XLNet, RoBERTa and BERT respectively. 
In contrast, RTE and MRPC tasks show substantial change (gain/drop) in the performance with layer-dropping when compared with using the full model (see BERT and RoBERTa $0/12$,$2/12$,$4/12$ results). This is due to the small size of the dev sets, 408 and 277 instances for MRPC and RTE respectively. A few right and wrong predictions cause a large variation in the overall score. We use McNemar’s test at p=value=0.05, and we found these differences, such as $5.05$ points drop in the performance of BERT for RTE, statistically insignificant. 

\textbf{Dropping top two layers for RoBERTa resulted in better performance and stability.}
Interestingly, in several cases for RoBERTa, %the drop of 
dropping two layers resulted in \emph{better} performance than using the full model. 
%particularly when dropping two layers, resulted in \emph{better} performance than when using the full model. 
Moreover, we observed that layer-dropping resulted in stable runs and was less prone to initialization seed and batch size. We used default settings for all the model and did not investigate the effect of parameter optimization on the performance of the pre-trained and reduced models to have comparable results.
%Note that we did not investigate the effect of parameter optimization on the performance of the pre-trained models and the reduced models. In order to have comparable results, we used default settings for all models. % in order to have comparable results.

%\textbf{One-third of a distilled model can also be pruned successfully.}
\textbf{A distilled model can also be pruned successfully.}
%\nd{Either be specific which distilled model you are talking about or better make the finding generic that we can also prune distilled models effectively using our approach} 
We observed a similar trend, %of 
dropping layers %from 
in DistilBERT compared to BERT model. %Here, 
%DistilBERT serves as an interesting case where 
It is interesting to see that an already distilled version of the model can be further pruned
% a model %is 
% designed to be a compressed version of a larger model, %but we can still remove one-third of its layers 
% can still be pruned further
by a third, with an average loss of $0.75$ points only. However, dropping half of %the DistilBERT 
its layers drastically degrades the performance on several tasks. \citet{schwartz2020right} also showed that pruning is orthogonal to model distillation. 

%We further extend our layer-dropping experiments to DistilBERT in order to probe whether a distilled model is also over-parameterized and does it  exhibit similar behavior to layer-dropping compared to their larger model counterpart.  
%Figure~\ref{fig:average_results_distilbert} presents the average results for DistilBERT. We see similar trends: the top strategy is most consistent, Odd-alternate dropping of single layer improved the performance by $0.22$ points over no dropping. %DistilBERT serves as an interesting case where a model is designed to be a compressed version of a larger model, but we can still drop one-third of its layers with an average loss of $0.75$ points only. 
%The consistency of observations across a large model and a distilled model show that the behavior of dropping layers is general to all models and even if a model is designed to be a compressed version of a large model, one can still drop one-third of its layers with respectable drop in performance. %average loss of $0.75$ points only. 

%We further investigate 
%\textbf{How many layers are strictly necessary for each individual task}.
%\textbf{Certain tasks require as few as 3 layers to maintain within 1\% performance threshold.}

%To summarize, 
%%for 12-layer models 
%we showed that while one can drop one-third to half top layers of the model, a task-specific optimization of layer-dropping results in a better balance between performance degradation and model size. 

\subsection{Memory and Speed Comparison}
Dropping layers reduces the number of parameters in the network, significantly speeding up the task-specific fine-tuning and the inference time. Table~\ref{tab:paramters_comparison} compares the number of parameters, and the speed up in the fine-tuning and decoding time, versus the loss in performance. We see that dropping top half of the layers of the network, reduced the number of parameters by $40\%$, speeding up fine-tuning and inference by $50\%$ with average performance loss between $0.89$--$2.91$ points. The results for RoBERTa are even remarkable; as with all the memory and speed improvements, the average performance dropped by only $0.89$ points. Dropping 4 layers (which %still 
gives a speed-up of $33\%$), RoBERTa achieved a performance close to dropping no layers. XLNet also showed robustness to the drop of top 4 layers and the performance dropped by only $0.23$ points. 
It is worth noting that a better trade-off between computational efficiency and loss in performance can be achieved by optimizing for a specific task. For example %, Table~\ref{tab:performance_results} shows that 
QQP maintained %a 
performance within $1\%$ on XLNet when $9$ layers were dropped (See Table~\ref{tab:performance_results}). This corresponds to $60\%$ reduction in the number of parameters and $80\%$ reduction in terms of inference time.

% 0 to 11 is top
\begin{table}[t]
\centering
\footnotesize
\begin{tabular}{c|c|lcc}
\toprule
Drop. & Loss & Param. & Fine-tuning & Inference \\
     &      &        &    speedup  &  seconds \\
\midrule
\multicolumn{5}{c}{BERT $\|$ RoBERTa} \\
$0/12$ & 0.00 \hphantom{-} $\|$ \hphantom{-}0.00 & 110M & 1.00 & - \\
$2/12$ & 1.33 \hphantom{-} $\|$ -0.42 & 94M & 1.24 & 17\% $\downarrow$ \\ 
$4/12$ & 2.00 \hphantom{-} $\|$ \hphantom{-}0.01 & 80M & 1.48 & 33\% $\downarrow$ \\
$6/12$ & 2.91 \hphantom{-} $\|$ \hphantom{-}0.89 & 66M & 1.94 &  50\% $\downarrow$ \\
% \midrule
% \multicolumn{5}{c}{RoBERTa} \\
% $0/12$ & 0.00 & 110M & 1.00 & - \\
% $2/12$ & -0.42 & 94M & 1.19 & 17\% $\downarrow$ \\
% $4/12$ & 0.01 & 80M & 1.48 & 33\% $\downarrow$ \\
% $6/12$ & 0.89 & 66M & 1.94 & 50\% $\downarrow$ \\
\midrule
\multicolumn{5}{c}{XLNET} \\
$0/12$ & 0.00 & 116M & 1.00 & - \\
$2/12$ & 0.54 & 101M & 1.20 & 16\% $\downarrow$ \\
$4/12$ & 0.23 & 86M & 1.49 & 32\% $\downarrow$ \\
$6/12$ & 1.81 & 71M & 1.96 & 49\% $\downarrow$ \\
% \midrule
% \multicolumn{5}{c}{DistilBERT} \\
% $0/6$ & 0.00    & 66M & 1.00 & - \\
% $1/6$ & 0.49 & 59M & 1.19 & 17\% $\downarrow$ \\
% $2/6$ & 0.75 & 52M & 1.48 & 33\% $\downarrow$ \\
% $3/6$ & 3.28 & 45M & 1.94 & 50\% $\downarrow$ \\
\bottomrule
\end{tabular}
\caption{\label{tab:paramters_comparison}Comparing the number of parameters (Param.), the speed up in the fine-tuning step, and the inference time for different models. \emph{Fine-tuning speedup} shows how many times the model speeds up compared to the original network. We report inference time on the QQP devset consisting of 40.4k instances with a batch size of 32.}
\end{table}

% % 0 to 11 is top
% \begin{table}[t]
% \centering
% \footnotesize
% \begin{tabular}{c|l|lcc}
% \toprule
% Drop. & Loss & Param. & Fine-tuning & Inference \\
%      &      &        &    speedup  &  seconds \\
% \midrule
% \multicolumn{5}{c}{BERT} \\
% $0/12$ & 0.00 & 110M & 1.00 & - \\
% $2/12$ & 1.33 & 94M & 1.24 & 17\% $\downarrow$ \\ 
% $4/12$ & 2.00 & 80M & 1.48 & 33\% $\downarrow$ \\
% $6/12$ & 2.91 & 66M & 1.94 &  50\% $\downarrow$ \\
% \midrule
% \multicolumn{5}{c}{RoBERTa} \\
% $0/12$ & 0.00 & 110M & 1.00 & - \\
% $2/12$ & -0.42 & 94M & 1.19 & 17\% $\downarrow$ \\
% $4/12$ & 0.01 & 80M & 1.48 & 33\% $\downarrow$ \\
% $6/12$ & 0.89 & 66M & 1.94 & 50\% $\downarrow$ \\
% \midrule
% \multicolumn{5}{c}{XLNET} \\
% $0/12$ & 0.00 & 116M & 1.00 & - \\
% $2/12$ & 0.54 & 101M & 1.20 & 16\% $\downarrow$ \\
% $4/12$ & 0.23 & 86M & 1.49 & 32\% $\downarrow$ \\
% $6/12$ & 1.81 & 71M & 1.96 & 49\% $\downarrow$ \\
% % \midrule
% % \multicolumn{5}{c}{DistilBERT} \\
% % $0/6$ & 0.00    & 66M & 1.00 & - \\
% % $1/6$ & 0.49 & 59M & 1.19 & 17\% $\downarrow$ \\
% % $2/6$ & 0.75 & 52M & 1.48 & 33\% $\downarrow$ \\
% % $3/6$ & 3.28 & 45M & 1.94 & 50\% $\downarrow$ \\
% \bottomrule
% \end{tabular}
% \caption{\label{tab:paramters_comparison}Comparing the number of parameters (Param.), the speed up in the fine-tuning step, and the inference time for different models. \emph{Fine-tuning speedup} shows how many times the model speeds up compared to the original network. We report inference time on the QQP devset consisting of 40.4k instances with a batch size of 32.}
% \end{table}

\section{Discussion}
\label{sec:discussion}

Now we perform further analysis and discuss variations of our methodology. %However, 
We limit the results to 5 most stable tasks (SST-2, MNLI, QNLI, QQP, STS-B).

\subsection{Task-specific optimal number of layers to drop.}

The variation in the amount of loss for each task with the dropping of layers in Table~\ref{tab:all_tasks} suggests that the task-specific optimal number of layers would result in a better balance between the size of the pruned model and the loss in performance. In this section, we present the results of the optimal number of layers for each task.
%In order to find the optimal number of layers that can be dropped for a task while maintaining performance within a certain threshold, 
For these experiments, we split the standard development set into equal-sized hold-out set and dev set. We find the minimum number of layers required to maintain 1\%, 2\%, and 3\% performance on the dev set using our top-layer strategy and we verify that the findings generalize to the hold-out test. 
Table~\ref{tab:performance_results} shows the optimal number of layers on dev and the corresponding percentage of performance drop on the hold-out set (in parentheses). 
For most of the cases, the optimal number of layers found using the dev set aligns well with the hold-out set. For example, BERT QNLI with 1\% loss in performance showed that one layer can be dropped safely and this results in a loss of 0.84 points absolute compared to using the full model.

Overall, RoBERTa and XLNet showed most robustness towards the dropping of layers while maintaining performance threshold of 1\%. For example, QQP maintained performance within $1$ point even when the top $9$ and $8$ layers of XLNet and RoBERTa respectively were dropped. Essentially, the model consists of only three layers -- $\{1,2,3\}$. On the contrary, %the drop of 
dropping $9$ layers in BERT resulted in a loss of 3\% points for the QQP task. 

%On the other hand, the QNLI and the MNLI tasks are most sensitive to dropping layers across all models, reflecting that these are more general tasks that require most of the network to perform well. This result is inline with recent research that uses MNLI fine-tuned models as a base for other task-specific fine-tuning \cite{roberta}. 

\begin{table}[t]									
\centering									
\footnotesize									
\begin{tabular}{l|ccccc}									
\toprule
 & SST-2 & MNLI & QNLI & QQP & STS-B \\		
\midrule
\multicolumn{6}{c}{1\% Loss Threshold}  \\
BERT & 7(1.6) & 3(1.04) & 1(0.84) & 6(0.75) & 7(1.16) \\
RoBERTa & 4(0.00) & 4(0.20) & 5(0.87) & 8(0.77) & 5(1.22) \\
XLNet & 8(1.38) & 5(1.22) & 4(0.51) & 9(0.60) & 7(0.05) \\
\midrule
\multicolumn{6}{c}{2\% Loss Threshold}  \\
BERT & 7(1.60) & 5(1.26) & 3(1.68) & 8(1.60) & 7(1.16) \\
RoBERTa & 4(0.00) & 5(1.26) & 6(1.42) & 9(1.51) & 6(2.31) \\
XLNet & 8(1.38) & 5(1.22) & 6(1.46) & 9(0.60) & 8(1.22) \\
\midrule
\multicolumn{6}{c}{3\% Loss Threshold}  \\
BERT & 8(2.06) & 6(2.42) & 5(2.60) & 9(2.27) & 8(2.61)  \\
RoBERTa & 5(0.69) & 6(2.73) & 7(2.37) & 10(3.21) & 7(3.00)  \\
XLNet & 8(1.38) & 6(1.55) & 7(1.61) & 9(0.60) & 9(2.46)  \\
%\midrule 
% \multicolumn{6}{c}{DistilBERT}  \\
% 1\% & 2 & 0 & 1 & 3 & 2 \\
% 2\% & 2 & 1 & 1 & 3 & 3 \\
% 3\% & 3 & 2 & 2 & 4 & 3  \\
\bottomrule
\end{tabular}
\caption{Number of layers dropped from the network while maintaining performance within a pre-defined threshold. The numbers outside brackets are the optimal number of layers found using the dev set and the numbers within brackets report the performance loss on the hold-out set. For example in 7(1.6), 7 are the optimal number of layers that can be dropped based on the dev set and 1.6 is the performance loss when 7 layers are dropped on the hold-out set. 
}
\label{tab:performance_results}						
\end{table}

\subsection{Comparing Pre-trained Models}
\begin{figure}[t]
\centering
 \includegraphics[width=0.6\linewidth]{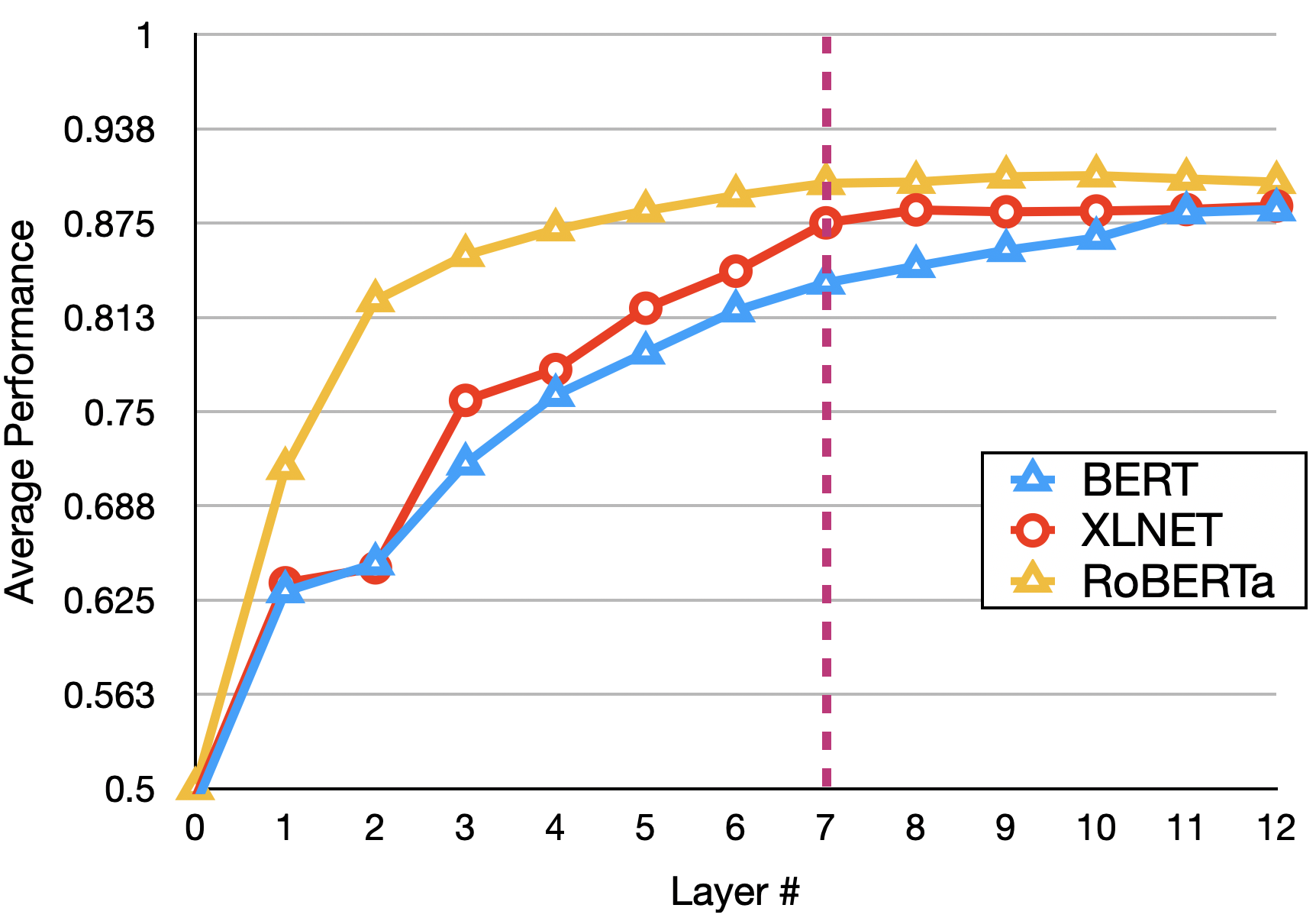}
%\vspace{-3pt}
%   \vspace{-2mm}
\caption{Average layer-wise classification results.}
%\vspace{-3pt}
\label{fig:layerwise_graph}
%\vspace{-4mm}
\end{figure}

Our pruning strategies illuminate model-specific peculiarities that help us in comparing and understanding the learning dynamics of these models. 
\textbf{RoBERTa and XLNet learn task-specific knowledge earlier in the network compared to BERT}. Figure \ref{fig:layerwise_graph} shows the average layer-wise performance of each model. RoBERTa learns task-level information much earlier in the model (see the steep slope of the yellow line for lower layers). Although XLNet starts similar to BERT but in the lower-middle layers, it learns the task information relatively faster than BERT. For both RoBERTa and XLNet, the performance matures close to the $7^{th}$ layer of the model while BERT improves with every higher layer until the $11^{th}$ layer. Since XLNet and RoBERTa mature much earlier in the network, %(\emph{i})~RoBERTa and XLNet learn task-specific knowledge at much lower layers compared to BERT, (\emph{ii})~the last layers of RoBERTa and XLNet 
this suggests that top layers in these networks might be redundant for downstream tasks and are a good candidate for dropping in exchange for a small loss in performance. This observation is in line with the results presented in Table~\ref{tab:all_tasks}. For example, we showed that the drop of top two layers of RoBERTa resulted in either marginal drop in performance or improvement in performance. 

The difference between the learning dynamics of BERT and RoBERTa encourages further investigation into what caused RoBERTa to learn task-specific knowledge earlier in the network. Is it because of the large amount of training data used for RoBERTa or because of better pre-training procedures such as dynamic masking, and exclusion of next sentence prediction loss? Does early learning of task-specific knowledge as in XLNet and RoBERTa reflect towards a better and robust pre-trained model? Answering these questions is important for improving the design of pre-trained models and require future exploration.

%\section{Practical Benefits of Layer-dropping}
%\label{sec:practical}
%The layer-dropping strategies offer an effective and low-cost method to reduce the size of pre-trained models for practical usages. Here, we discuss the speed up achieved in comparison with the drop in performance. Later, we put the performance of pruned models using layer-dropping into the perspective of distilled models built using knowledge distillation.

\subsection{Pruning the ALBERT Model}
ALBERT is 
based on the 
%designed to be memory efficient using 
cross-layer parameter sharing. 
%\sout{thus} \sout{therefore the number of parameters do not grow with the increase in the depth of the network. We apply top-layer dropping to ALBERT.} 
Because of this, our layer dropping strategies do not save any memory as opposed to using BERT and other transformers. However, it still makes the inference faster by speeding up the forward pass. Table~\ref{tab:albert} presents the results on five GLUE tasks. Interestingly, dropping the top-6 layers did not result in drastic degradation of the model performance and, in some cases, the results even improved compared to using the baseline model. For example, in the case of SST-2, the performance of a 6-layered model is 90.14, which is 0.35 points absolute better than the baseline. 
Compared to the 6-layered BERT model (Table~\ref{tab:all_tasks}), the drop in the performance of ALBERT-6 is relatively small. We hypothesize that the parameter sharing in the case of ALBERT make the model learn much richer representation in the shared contextualized layers of the model, which yields a model that is robust towards layer-dropping. These results are encouraging and show that the model that was designed to be space-efficient can be further improved towards run-time efficiency by simply pruning some of its layers.

\begin{table}[t]									
\centering
\begin{tabular}{l|lllll%|ll
}	

\toprule									
									
Drop & SST-2 & MNLI & QNLI & QQP & STS-B  \\ % & RTE & MRPC \\		
\midrule						
%\multicolumn{6}{c}{ALBERT}  \\		
$0/12$ & 89.79 & 83.39 & \textbf{90.24} & \textbf{90.29} & \textbf{89.61} \\ % & 67.51 & 74.02 \\
$2/12$ & 91.40 & \textbf{83.82} & 89.55 & 89.64 & 89.54 \\ % & 69.31 ({\color{blue}1.80$\downarrow$}) & 88.48 ({\color{red}0.00$\downarrow$}) \\
$4/12$ & \textbf{91.63}  & 82.73 & \textbf{90.24} & 88.51 & 87.00 \\ % & 66.06 ({\color{blue}1.45$\uparrow$}) & 85.29 ({\color{red}0.00$\downarrow$}) \\
$6/12$ & 90.14 & 81.64 & 89.11 & 90.08 & 88.21 \\ %& 53.79 ({\color{red}13.72$\downarrow$}) & 83.58 ({\color{red}0.00$\downarrow$}) \\

% $0/12$ & 89.79 & 83.39 & 90.24 & 90.29 & 89.61 \\ % & 67.51 & 74.02 \\
% $2/12$ & 91.40 ({\color{blue}1.61}) & 83.82 ({\color{blue}0.43$\downarrow$}) & 89.55 ({\color{red}0.69$\downarrow$}) & 89.64 ({\color{red}0.65$\downarrow$}) & 89.54 ({\color{red}0.07$\downarrow$}) \\ % & 69.31 ({\color{blue}1.80$\downarrow$}) & 88.48 ({\color{red}0.00$\downarrow$}) \\
% $4/12$ & 91.63 ({\color{blue}1.84$\downarrow$}) & 82.73 ({\color{red}0.66$\downarrow$}) & 90.24 ({\color{blue}0.00}) & 88.51 ({\color{red}1.78$\downarrow$}) & 87.00 ({\color{red}2.61$\downarrow$}) \\ % & 66.06 ({\color{blue}1.45$\uparrow$}) & 85.29 ({\color{red}0.00$\downarrow$}) \\
% $6/12$ & 90.14 ({\color{blue}0.35$\downarrow$}) & 81.64 ({\color{red}1.75$\downarrow$}) & 89.11 ({\color{red}1.13$\downarrow$}) & 90.08 ({\color{red}0.21$\downarrow$}) & 88.21 ({\color{red}1.40$\downarrow$})  \\ %& 53.79 ({\color{red}13.72$\downarrow$}) & 83.58 ({\color{red}0.00$\downarrow$}) \\
\bottomrule
\end{tabular}

\caption{ALBERT: task-wise performance for the top-layer dropping strategy using the official GLUE dev-sets. \textbf{Drop} shows the number of layer dropped/the total layers in the model.}
\label{tab:albert}									
\end{table}

%\paragraph{Pruned BERT/XLNet vs. DistilBERT}
\subsection{Comparing against Distilled Models}
We now compare the performance of our pruned models when applying the top-layer dropping strategy to distilled and pruned models built using various sophisticated architectures and training procedures. In particular, we compare to previous work~\cite{sanh2019distilbert,turc2019wellread,tang2019distiltaskspecific} that used KD to build 6-layered distilled models. 
More specifically, we present the result of the following distilled models; Vanilla-KD -- a distilled model built using the original KD method~\cite{Hinton06}, BERT-PKD~\cite{sun-etal-2019-patient} -- patient knowledge distillation method that encourages a student model to learn from various layers of the teacher model, and BERT-TH -- a theseus compression method that gradually distill layers of a large model. 
Additionally, we compare with the pruned RoBERTa model of \cite{fan2019reducing} that used layer-level dropout during training of a pre-trained model and showed that it enables robust dropping of layers at test time. We also compare to the greedy layer pruning method~\cite{greedy_layer_pruning}, which creates task-specific smaller-size models by dropping layers in a greedy fashion.
All these models are identical in size to our smaller models obtained by dropping the top-6 layers in BERT and RoBERTa. We refer to them as \textit{BERT-6} and \textit{RoBERTa-6}. %, respectively.
%
%Dropping half of the encoder layers of a 12-layer model resulted in a 6-layer reduced model. We refer to them as \textit{BERT-6/RoBERTa-6}. \citet{sanh2019distilbert,turc2019wellread,tang2019distiltaskspecific} and others have used knowledge distillation (KD) to build 6-layered distilled models of BERT. 
%Most of these models\footnote{Some models have different layer sizes.} are identical in size to our reduced models, BERT-6 and RoBERTa-6. %smaller model obtained by dropping top 6 layers of BERT. 
%We refer to our 6-layered model achieved using the top-strategy as \textit{BERT-6/RoBERTa-6}. 
%In this section, we compare the performance of our reduced models with several models built using KD. 
%Table \ref{tab:bertvsdistilbert} compares the results.\footnote{We did not intend to compare with an exhaustive list of task-specific distilled models and show the results of a few of them for comparison purposes.} % of several 6-layered models with our BERT-6 and RoBERTa-6 models. 
%Only DistilBERT~\cite{sanh2019distilbert} is directly comparable to BERT-6 since it is a general-purpose distilled model. Rest of the 6-layered distilled models are task-specific models. 
%The results of distilled models are copied from~\cite{Xu2020BERTofTheseusCB}. 
Table \ref{tab:bertvsdistilbert} compares the results.\footnote{%We did not intend to compare with an exhaustive list of task-specific distilled models and show the results of a few of them for comparison purposes. 
There is an exhaustive list of task-specific distilled models but we show the results for a few for comparison.}

%After dropping $6$ layers from BERT, and XLNet, the size of the resulting networks is identical to DistilBERT. We refer to our 6-layered model as BERT-6 later on. We compile the results for these models in Table~\ref{tab:bertvsdistilbert}.\footnote{We %intentionally kept 
%left out MRPC and RTE of the analysis intentionally, as we found them extremely unstable for all runs including the ones that use the full network. This may be due to the small size of their training data: only 3.6k and 2.5k instances for MRPC and RTE respectively. The results are presented in Table~\ref{tab:all_tasks}.} 

\textbf{Our pruned models (BERT-6 and RoBERTa-6) showed competitive performance compared to their distilled versions built using KD.}
%
%Our BERT-6 model showed competitive performance to DistilBERT. 
%%Compared to task-specific models, the BERT-6 model outperformed on the QQP and STS-B tasks while it is only lower by 1.17 to 1.55 points to the best task-specific model. 
This result is quite surprising, given that our pruned models do not require any additional training, while building a distilled model using KD requires training from scratch, which is a time consuming and computation expensive process. 
%Our top strategy performs consistently well to different model types including distilled models and to various types of sequence classification tasks. 
The top layer-dropping works consistently for all model types including distilled models and a large set of language understanding tasks. Moreover, our setup offers the flexibility to choose different sizes of the model based on the computational requirements and the specifics of a downstream task. 
%The top-layer dropping strategy is similar to the baselines used by~\cite{sun-etal-2019-patient} and \cite{Xu2020BERTofTheseusCB} where they use first six layers of the BERT model to initialize a student model and apply task-specific fine-tuning. However, the results reported in their papers are much lower than ours. We provide a comprehensive evaluation of the top-layer dropping strategy on four pre-trained models to prove its  efficacy in reducing the size of the network while maintaining most of the performance. 
The result of preserving bottom layers of the model suggests selective compression applied to pre-trained models. For example, in KD while combining information from various layers of the large model, it is advisable to preserve the bottom layers and distilled the top layers. Similarly, pruning methods such as weight and attention-head pruning, and quantization can be aggressively applied to top layers of the models while preserving the bottom layers. % untouched. 

\begin{table}[t]
\centering									
\footnotesize
%\resizebox{\columnwidth}{!}{
\begin{tabular}{l|l|ccccc}						
\toprule
No. & Model & SST-2 & MNLI & QNLI & QQP & STS-B \\	
\midrule						
%\multicolumn{6}{c}{General-purpose BERT distilled models} \\

%\midrule
%\multicolumn{6}{c}{Task-specific BERT distilled models} \\
1. & Vanilla-KD & 90.50 & 80.10 & 88.00 & 88.10 & 84.90 \\
2. & BERT-PKD & 91.30 & 81.30 & 88.40 & 88.40 & 86.20 \\
3. & BERT-TH & \textbf{91.80} & \textbf{82.10} & 88.80 & 88.80 & 87.80 \\
4.   & GLP$_6$ & 91.20 & 81.30 & 87.60 & 86.80 & 87.60 \\
5. & DistilBERT & 90.37 & 81.78 & \textbf{88.98} & \textbf{90.40} & 87.14 \\
6. & BERT-6 & 90.25 & 81.13 & 87.63 & \textbf{90.35} & \textbf{88.45} \\
\midrule
7. & Fan et al. RoBERTa-6 & 92.50 & 82.90 & 89.40 & - & - \\
8.   & GLP$_6$ & 92.00 & \textbf{85.60} & \textbf{90.80} & 87.80 & 86.60 \\
9. & DistilRoBERTa & \textbf{92.50} & 84.00 & \textbf{90.80} & 89.40 & 88.30  \\
10. & RoBERTa-6 & 91.97 & 84.44 & 90.00 & \textbf{90.91} & \textbf{88.92} \\
\bottomrule
\end{tabular}
%}
\caption{
%Task-wise performance for d
Comparing 6-layered BERT and RoBERTa models. Results of Vanilla-KD, BERT-PKD and BERT-TH are taken from~\citet{Xu2020BERTofTheseusCB}. Fan et al. results and GLP$_6$ are taken from %their papers 
\cite{fan2019reducing,greedy_layer_pruning}. BERT-6 and RoBERTa-6 represent our models achieved by pruning top 6 layers. %Vanilla-KD \cite{Hinton06}, BERT-PKD (Patient KD~\cite{sun-etal-2019-patient}, BERT-TH (BERT of Theseus~\cite{Xu2020BERTofTheseusCB}).
}

\label{tab:bertvsdistilbert}									
\end{table}									

\textbf{Our RoBERTa-6 has comparable results to the 6-layer pruned model trained using LayerDrop and Greedy layer pruning.}
\citet{fan2019reducing} used layer-level dropout during training of a pre-trained model and showed that it enables robust dropping of layers at test time. Similar to us, they directly pruned top 6-layers of their large model and fine-tuned it for specific tasks. Table \ref{tab:bertvsdistilbert} (row 7 and 10) compares top-layer dropping using their model and the original RoBERTa model. On two out of three tasks, dropping top-layers from the original RoBERTa model outperformed training a new model using LayerDrop. 
This shows that the current models are already robust %to dropping top-layers and 
and the top-layer dropping strategy can be directly applied to the available pre-trained models. Similarly, we found that despite optimizing the model towards a downstream GLUE task, the greedy layer pruning (GLP$_6$) did not show a clear advantage over our 6-layered model. For example, compared to BERT (rows 4 and 6), our BERT-6 model yields better or comparable performance to GLP$_6$ on the QQP, STS-B, MNLI and QNLI tasks, and performs worse only on the SST-2 task.

% \begin{table}[t]
% \centering
% \resizebox{\columnwidth}{!}{
% \begin{tabular}{l|cccc}						
% \toprule
% Model & SST-2 & MNLI & QNLI & MRPC \\	
% \midrule						
% RoBERTa+LayerDrop & \textbf{92.50} & 82.90 & 89.40 & \textbf{85.30} \\
% RoBERTa-6 & 91.97 & \textbf{84.44} & \textbf{90.91} & 82.84 \\
% \bottomrule
% \end{tabular}
% }
% \vspace{-3pt}
% \caption{Results of applying top-layer dropping strategy on the RoBERTa model built with and without LayerDrop. RoBERTa+LayerDrop results are taken from~\cite{fan2019reducing}.}
% \vspace{-3pt}
% \label{tab:roberta_layerdrop}									
% \end{table}	

\subsection{Layer-Dropping using Fine-tuned Models}
%One of the advantages of our dropping strategies is that they are directly applied to the pre-trained models, i.e.,~we avoid the need to optimize our strategies for each task. However, it is possible that dropping from a fine-tuned model may result in better performance. 
Here, we question whether dropping layers from a fine-tuned model is more effective than dropping them from a base model? %dropping layers from a per-trained model, are we losing any useful information which is essential for the downstream task? 
%
%applying layer-dropping on fine-tuned models would result in different behaviour?  
To answer this, we first fine-tune the model, drop the layers, and then fine-tune the reduced model again. Table~\ref{tab:xlnet_ft_drop} presents the results on BERT and XLNet. We found this setup to be comparable to dropping layers directly from the pre-trained model in most of the cases. This shows that dropping top layers directly from a pre-trained model does not lose any critical information which was essential for a specific task. However, we think that pruning a fine-tuned model may lose task-specific information %since %after 
because %during fine-tuning, %the 
the model is optimized for the task. Dropping layers may have severe effect. This is reflected in some of the results of BERT-6. %inforbefore dropping, the chance of losing what the model has optimized for task are high.\todo{(P) This sentence is broken...} 
%Other disadvantages of this method are that (\emph{i})~it builds task-specific reduced models instead of one general reduced model that can be used for task-specific fine-tuning, and (\emph{ii})~it requires running fine-tuning twice for each task, which is a time-consuming process. 

\begin{table}[t]									
\centering									
%\resizebox{\columnwidth}{!}{							
\footnotesize
\begin{tabular}{l|ccccc}									
\toprule
Model & SST-2 & MNLI & QNLI & QQP & STS-B \\		
\midrule						
BERT-$6$ & \textbf{92.25} & \textbf{81.13} & \textbf{87.63} & 90.35 & \textbf{88.45} \\
BERT-FT-$6$ & 90.02 & 80.85 & 87.24 & 90.34 & 88.16 \\
\midrule						
XLNet-$6$ & 92.20 & 83.48 & \textbf{88.03} & 90.62 & 87.45  \\
XLNet-FT-$6$ & \textbf{92.43} & \textbf{83.75} & 86.80 & \textbf{90.77} & \textbf{87.60} \\
\bottomrule
\end{tabular}
%}
%\vspace{-2pt}
%  \vspace{-1mm}
\caption{Layer-dropping using task-specific models. XLNet-FT-$6$ first fine-tunes the pre-trained model, %and then it 
removes the layers and performs fine-tuning again.}	
%\vspace{-6pt}
\label{tab:xlnet_ft_drop}									
\end{table}

%\subsection{Gradual Dropping}
\paragraph{Gradual Dropping:} In another attempt 
%Similarly to layer-dropping using fine-tuned model, here we want 
to preserve the model's performance during the dropping process, %instead of dropping all layers together, 
we iteratively drop one layer after every two epochs of the fine-tuning process. This did not yield any improvement over dropping layers directly from the 
%pre-trained 
model.

\section{Conclusion}
% \vspace{-6pt}
\label{sec:conclusion}
We proposed %several 
strategies to drop layers in pre-trained models and analyzed the model behavior on downstream tasks. We conducted experiments using a variety of pre-trained models and using a diverse set of natural language understanding tasks and showed that one can reduce the model size by up to 40\%, while maintaining up to 98\% of their original performance on downstream tasks. Our pruned models performed on par with distilled models building using knowledge distillation. However, unlike distilled models, our approach does not require re-training, is applicable to a large set of pre-trained models including distilled models, and provides the flexibility to balance the trade-off between accuracy and model size. 
Moreover, we made several interesting observations such as, i) the lower layers are most critical to maintain downstream task performance, ii) certain downstream tasks require as few as only 3 layers out of 12 layers to maintain within 1\% performance threshold, iii) networks trained using different objective functions have different learning patterns e.g. XLNet and RoBERTa learns task-specific information much earlier in the network compared to BERT. %In addition to the analysis, our setup is of practical importance. We showed that dropping half of the encoder layers maintains up to \textcolor{red}{98.2\% of the original accuracy while reducing the model size by 40\%. } We showed that the pruned models using layer-dropping have competitive performance to distilled models built using knowledge distillation. In addition, 
\bibliographystyle{elsarticle-num-names}
\bibliography{acl2020}

\end{document}